\documentclass[journal]{IEEEtran}
\usepackage{graphicx}
\usepackage{comment}
\usepackage{amsmath,amssymb} 
\usepackage{color}
\usepackage{mathtools}
\usepackage{pifont}

\usepackage{bm}
\usepackage{caption}
\usepackage{makecell}
\newcommand{\red}{\textcolor{black}}%

\newcommand{\revised}{\textcolor{black}}%

\newcommand{\etal}{\textit{et al. }}

\ifCLASSINFOpdf
\else
\fi
\begin{document}
%
\title{Clustering Aided Weakly Supervised Training to Detect Anomalous Events in Surveillance Videos}

%
%
%
\author{Muhammad Zaigham Zaheer,
        Arif Mahmood,
        Marcella Astrid,
        and~Seung-Ik Lee
        
\IEEEcompsocitemizethanks{\IEEEcompsocthanksitem M. Zaigham Zaheer, Marcella Astrid and Seung-Ik Lee are with the University of Science and Technology and Electronics and Telecommunications Research Institute, Daejeon, Korea.
\protect\\
E-mail: see sites.google.com/view/cvml-ust/members 
\IEEEcompsocthanksitem Arif Mahmood is with the Information Technology University, Pakistan.}
\thanks{Manuscript received April 19, 2005; revised August 26, 2015.}}

%
%

\markboth{Journal of \LaTeX\ Class Files,~Vol.~14, No.~8, August~2015}%
{Shell \MakeLowercase{\textit{et al.}}: Bare Demo of IEEEtran.cls for IEEE Journals}
%



\maketitle

\begin{abstract}

Formulating learning systems for the detection of real-world anomalous events using only video-level labels is a challenging task mainly due to the presence of noisy labels as well as the rare occurrence of anomalous events in the training data. We propose a weakly supervised anomaly detection system which has multiple contributions including a random batch selection mechanism to reduce inter-batch correlation and a normalcy suppression block which learns to minimize anomaly scores over normal regions of a video by utilizing the overall information available in a training batch. In addition,  a clustering loss block is proposed to mitigate the label noise and to improve the representation learning for the anomalous and normal regions. This block encourages the backbone network to produce two distinct feature clusters representing normal and anomalous events. Extensive analysis of the proposed approach is provided using three popular anomaly detection datasets including UCF-Crime, ShanghaiTech, and UCSD Ped2.
The experiments demonstrate a superior anomaly detection capability of our approach.


\end{abstract}


%
\IEEEpeerreviewmaketitle

\section{Introduction}

Anomalous event detection is an important problem which has numerous real-world applications including traffic management \cite{doshi2020fast,wang2019anomaly}, crowd management \cite{wu2010chaotic}, health and medicine \cite{wei2018anomaly}, cyber security \cite{gong2019memorizing}, and surveillance security systems \cite{mohammadi2016angry,sultani2010abnormal,kamijo2000traffic,sultani2018real,shanghaiTech2017}. Anomalies may be attributed as deviations from normal behaviors, activities, appearances or patterns. Therefore, a commonly used approach towards anomaly detection is the training of a one-class classifier which may learn frequently occurring patterns and appearances using only normal training samples \cite{sultani2018real,liu2018future,zhang2016video,luo2017revisit,xia2015learning,hinami2017joint,sabokrou2017deep_novelty,smeureanu2017deep,ravanbakhsh2018plug,ravanbakhsh2017abnormal,hasan2016anomaly}. Any example deviating from the learned normal representations is then considered as anomaly. A drawback of one-class classification methods is the limited availability of the training data not capturing all the normalcy variations \cite{chandola2009anomaly}. 
Hence, an occurrence of an unseen normal activity may significantly deviate from the learned representations and labeled as anomaly, resulting in an increased number of false alarms {\cite{hasan2016anomaly}}. Recently, weakly supervised learning algorithms \cite{liu2019completeness,liu2019weakly}, \revised{\cite{gong2020centroid}}, \cite{yu2019temporal,narayan20193c,shou2018autoloc,wang2017untrimmednets} have gained popularity, raising a different approach to anomaly detection by training a binary classifier using normal and weakly labeled anomalous  data \cite{sultani2018real,zhong2019graph}. Specifically, for the case of video based anomalous event detection, a video is labeled  as anomalous if \textit{some} of its contents are anomalous and labeled as normal if \textit{all} of its contents are normal. It means that a video labeled as anomalous may also contain numerous normal frames. Weakly supervised algorithms eliminate the need of temporal or spatial annotations, considerably reducing the costly efforts of obtaining manual fine-grained annotations of the training samples.

Recently, Sultani \etal formulated the weakly supervised anomaly detection task as Multiple Instance Learning (MIL) problem \cite{sultani2018real}. They considered a video as a bag of segments such that each segment consists of several consecutive frames. The training is then performed using video-level annotations by computing a ranking loss between the two top-scoring segments, one from the anomalous bag and the other from the normal bag.
Despite being an elegant approach, it represents each video using the same number of segments throughout the dataset which may result in information loss. Specifically, in a long video, if abnormal events occur in a short temporal range, such rigid formulation may make the detection significantly difficult.
More recently, Zhong \etal proposed learning of weakly supervised anomaly detection under noisy labels, where the noise refers to the normal contents within anomalous videos \cite{zhong2019graph}. Although their method demonstrates superior performance, it is prone to data correlation because of the training using consecutive batches from a video. In case of the videos recorded using stationary cameras, most of the frames consist of almost the same content, resulting in strong correlation. Several existing works have reported the deterioration of learning performance in deep networks due to the training data correlation \cite{bengio2012practical,mnih-atari-2013,mnih2015human}. 

In the current work, we propose to de-correlate the training data by using a batch based training architecture, where each batch consists of several temporally consecutive segments of a video. A lengthy video may be divided into several batches. At each training iteration, a batch is randomly selected across the full training dataset to eradicate inter-batch correlation. 
Note that the temporal consistency within a batch is still retained, which is necessary to carry out the proposed weakly supervised training.
Extensive evaluation of the proposed random batch selection method demonstrates that it significantly improves the performance compared to the previously proposed methods which use temporally correlated batches \cite{sultani2018real,zhong2019graph}.


Detecting anomalies in long untrimmed videos naturally raises the question of where we should look at, since in most cases, anomaly scenes have distinct characteristics than most of the normal data. Therefore, an attention mechanism may assist our model in detecting anomalous events. However, since conventional attention mechanisms are used to highlight the important features corresponding to the class annotations of the training data
\cite{chen2018attention,woo2018cbam,hu2018squeeze,wang2017residual}, their application is mostly limited to fully supervised scenarios.
\revised{Due to the weakly supervised nature of our approach and given the abundantly available normal segments in the training data, we pose the attention mechanism as suppressing the features that correspond to the normal events rather than highlighting them.} To this end, we propose a normalcy suppression mechanism operating over a full batch and learning to suppress the features corresponding to the normal contents. Our formulation is based on the intuition that the normal videos contain only normal contents and the anomalous videos contain some anomalies along with the normal contents.
Therefore, from the training point of view, learning to highlight the abnormal parts based on the relatively small amount of noisy data coming from anomalous videos is not feasible. \revised{In contrast, learning to suppress normalcy would be easier and well supported by the relatively large amount of noise-free normal data.}
Thus, for the case of an anomalous video input, the proposed suppression mechanism minimizes the impact of normal contents within that input thus indirectly highlighting the anomaly regions. In the case of an input video containing only normal contents, the suppression spans across the whole batch, thus forcing the backbone network to generate lower scores expected for the normal content. This way of suppression shows an improved performance than the conventional attention mechanism of highlighting, as explained further in Section \ref{sec:experiments}.

Motivated by the usage of clustering techniques for semi-supervised training \cite{kamnitsas2018semi,fogel2019clustering,shukla2018semi}, we also propose to incorporate an unsupervised clustering based loss. 
Since an anomalous labeled video may also contain normal segments and the anomaly detection being inherently a binary classification problem, we formulate a loss by grouping the segments of each video into two clusters. During consecutive training iterations, this loss encourages the network to maximize the distance between the two clusters for an anomalous labeled video. However, for the case of a normal video, as both clusters should belong to the normal class, the loss encourages the network to minimize the distance between these two clusters. In addition, the formulated loss also attempts to reduce the spread of each cluster in consecutive iterations. 
Consequently, the proposed clustering loss enforces the network to generate discriminative representations, enhancing the performance of our system towards anomaly detection. \revised{It may be noted that, although our overall proposed approach is demonstrated for anomaly detection applications, it may also be tweaked and extended to other computer vision problems such as action localization \cite{ma2020sf,zhao2021soda} and object localization \cite{zhang2020weakly,zhang2021weakly} in weakly supervised setting. }

The main contributions of the current work are as follows:
\begin{itemize}
  \item Using only video-level annotations, the proposed CLustering Assisted Weakly Supervised (CLAWS Net+) framework is trained to localize anomalous events in a weakly supervised manner.
  \item We reduce inter-batch correlation by using a simple yet effective random batch selection scheme to enhance the performance of the proposed framework.
 \item Our proposed normalcy suppression mechanism learns to suppress the features corresponding to the normal segments of an input by exploiting temporal information in a batch.
 \item We formulate a clustering based loss which enforces the network to minimize the distance between the clusters from a normal video and maximize it for the clusters from an anomalous video while simultaneously increasing the compactness of each cluster.
  \item Our proposed CLAWS Net+ framework demonstrates improved frame-level AUC performances of \red{84.16}\% on UCF-Crime {\cite{sultani2018real}}, \red{91.46}\% on ShanghaiTech \cite{shanghaiTech2017}, and \red{95.79}\% on UCSD-Ped2 \cite{chan2008ucsd} datasets, outperforming the existing state-of-the-art approaches {\cite{zhong2019graph,lu2013abnormal,hasan2016anomaly,sultani2018real}.}
\end{itemize}

A preliminary version of this work was recently presented in the European Conference on Computer Vision (ECCV) 2020 as CLAWS Net \cite{zaheer2020claws}. The current work, CLAWS Net+, is a substantial extension of the conference version. First, we extend the idea of utilizing clusters to assist the network training by reformulating the clustering based loss to incorporate cluster compactness. Second, we evaluate the performance of the system using two different feature extractors \cite{tran2015c3d,hara3dcnns}. Third, we extend the scope of our evaluations to three datasets by including UCSD-Ped2 in the experiments. Fourth, we present extended analysis of the proposed anomaly detection system and discuss various design choices.

The rest of the paper is organized as follows: Related work is discussed in Section II, Proposed CLAWS Net+ framework is described is Section III, Experiments follow in Section IV, and Conclusions are provided in Section V.

\section{Related Work}
In this section, we discuss the two popular categories of anomaly detection including one-class classification and weakly supervised binary classification. In addition to that, we review the methods using clustering as supervision and those using attention mechanisms to improve the performance.

\subsection{Anomaly Detection as One Class Classification (OCC)} One of the most popular paradigm for anomaly detection is to learn only normal representations and then at test time, data instances deviating from the learned normal behaviors are considered anomalous. In this category, the usage of handpicked features  \cite{medioni2001event_twostream34,basharat2008learning_realworld7,wang2014learning_realworld38,zhang2009learning_twostream53,piciarelli2008trajectory_twostream36} and deep features extracted using pre-trained models \cite{smeureanu2017deep,ravanbakhsh2017abnormal} have been experimented by several researchers. With the popularity of generative architectures, many researchers proposed to use generative networks that learn normalcy definitions in an unsupervised manner \cite{Gong_2019_ICCV,ren2015unsupervised,xu2015learning_denoise,ionescu2019objectcentric,Nguyen_2019_ICCV,nguyen2019hybrid,xu2017detecting_denoise,sabokrou2017deep_novelty,sabokrou2020deep}. These approaches rely on the assumption that a generative architecture will not be able to well-reconstruct the instances that are out of the learned distributions, thus may produce high reconstruction error for the case of anomalies. 
However, due to the limit of using only normal class data for training, it is difficult to ensure an effective classifier boundary that encloses normal data while excluding anomalies \cite{zaheer2020old}.
To mitigate this issue, some researchers recently proposed the idea of pseudo-supervised methods in which pseudo-anomaly instances are created using normal training data \cite{zaheer2020old,ionescu2019objectcentric}. Although this setting transforms the training to binary class problem, the overall training still  utilizes only normal training data hence falls under the category of OCC. In contrast to that, our current work deviates significantly from these methods as we do not use one-class training protocol. Instead, we utilize weakly-labeled anomalous and normal videos to train our architecture.

\subsection{Anomaly Detection as Weakly Supervised Learning}
\revised{This category of approaches utilizes either noisy or partial annotations to perform training on image datasets \cite{gong2020centroid}}, \cite{li2017learning,goldberger2016training,vahdat2017toward,patrini2017making,azadi2015auxiliary,natarajan2013learning,larsen1998design}. In these methods, either the loss correction is applied \cite{azadi2015auxiliary} or models are specifically trained to separate out the noisy labeled data \cite{li2017learning,vahdat2017toward}. \revised{In essence, the current work is different from these methods as we aim to handle video-level anomalies requiring temporally-ordered frame sequences}.

\revised{Recently, weakly-supervised action localization methods have also been proposed \cite{ma2020sf,zhao2021soda,liu2021weakly}.
While the underlying problem may be seen as a bit similar to temporal anomaly localization in surveillance videos, the challenges posed in these methods are fairly different. For example, \cite{ma2020sf} uses one-frame action category annotation to carry out the weakly-supervised training. Moreover, action recognition datasets generally assume frequent action frames across a particular category. In contrast, the anomalies are often attributed as rare events. Moreover, anomaly detection approaches identify anomalous vs normal without any further categorization of these events.}

Essentially, the closely related existing methods to ours are by Sultani \etal \cite{sultani2018real} and Zhong \etal \cite{zhong2019graph}, which also aim to perform anomaly detection using video-level annotations. Sultani et al. \cite{sultani2018real} formulated the weakly supervised training as Multiple Instance Learning (MIL) problem by considering a video as a compressed bag of segments. To carry out the training, a few top anomaly scoring segments from anomalous and normal bags are used to compute a ranking loss. In each training iteration, several pairs of such bags are utilized.
Zhong et al. \cite{zhong2019graph} employed graph convolutional neural networks to clean noisy labels in anomalous videos. Several consecutive training steps are performed using one complete training video. Therefore, it becomes difficult to avoid the correlation inherently present in the segments of a video. On the contrary, we propose a batch based training mechanism, where each batch comes from a different video.
In addition to that, we also propose a normalcy suppression mechanism which exploits the \textit{noise-free} normal training data to learn to suppress normal portions of a video. We also utilize unsupervised clustering for efficient noise cleaning which improves the anomaly detection performance of the proposed framework by encouraging it to produce distant clusters in the case of anomalous videos and closer clusters in the case of normal videos.


\subsection{Clustering as Supervision} The idea of using unsupervised clustering algorithms to assist the training of neural networks has become popular lately. Caron \etal \cite{caron2018deep} proposed the idea of using clustering labels as a supervisory signal to pre-train a deep network. Recently, Jain \etal \cite{jain2020actionbytes} proposed the idea of using pseudo labels generated from clustering to train an action classifier. These approaches follow the protocol of self-supervised learning where initial training is carried out using cluster-based pseudo labels and fine-tuning is carried out afterwards using the actual labels. \revised{Yuan et al. \cite{yuan2020self} have recently proposed a self supervision based tracking approach.} Another recent approach proposed by Zaheer \etal \cite{zaheer2020self} attempts to localize anomalous portions of a video by using the labels generated through clustering. However, the downside of using cluster-based labels directly as a supervision is that the performance relies heavily on the quality of the clustering algorithms. Therefore, without any specific supervision, convergence is not guaranteed. On the contrary, in our approach, we propose the utilization of unsupervised clustering algorithm to compute our formulated clustering loss that assists the end-to-end joint training of our complete framework.

\subsection{Normalcy Suppression} The normalcy suppression utilized in our architecture can be seen as a variant of attention {\cite{chen2018attention,woo2018cbam,hu2018squeeze,wang2017residual}}. \revised{However, due to rare occurrence of anomalies in real-world scenarios, we address the problem in terms of suppressing features, which is opposite to the conventional attention that highlights features {\cite{woo2018cbam,hu2018squeeze,wang2017residual}}. Specifically, we design the solution by keeping in mind the availability of \textit{noise-free} abundant normal video annotations.} Therefore, unlike the conventional attention utilizing a weighted linear combination of features, our approach performs an element-wise product of the features with suppression scores to suppress the features belonging to the normal portions of the input.

\begin{figure*}[t]
\begin{center}
   \includegraphics[width=1\linewidth]{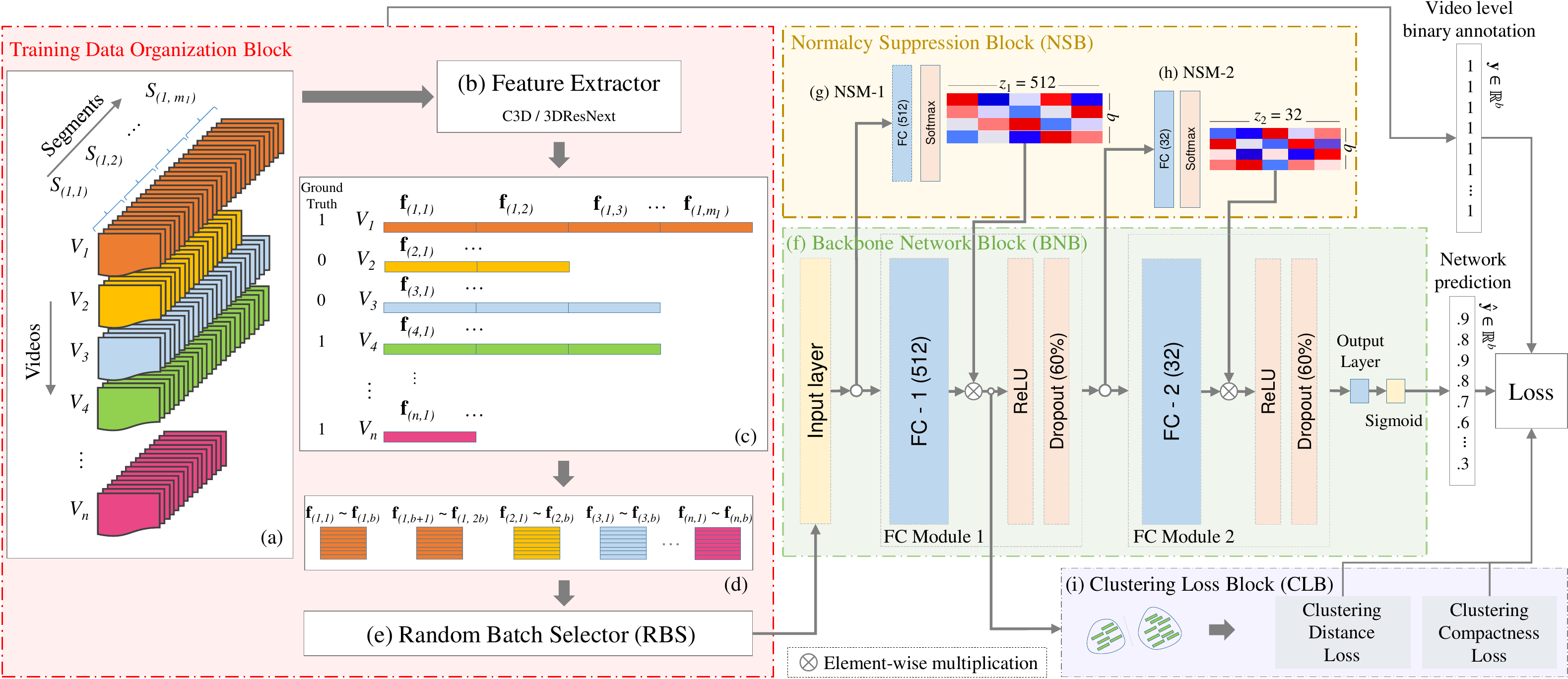}
\end{center}
   \caption{CLAWS Net+: The proposed weakly supervised anomaly detection framework using video-level labels. (a) Each input video is divided into equal length segments. (b) \& (c) Using each video segment, a feature vector is extracted. (d) Feature vectors are arranged into batches maintaining temporal order. (e) For training, batches are randomly selected. (f) Backbone network block consists of FC Module-1 \& 2. (g) \& (h) Normalcy suppression block consists of normalcy suppression modules, NSM-1 \& 2. (i) Clustering loss block in which a loss is computed using two clusters created in an unsupervised fashion.
   }
\label{fig:architecture}
\end{figure*}
\section{Proposed CLAWS Net+ Architecture}
In this section, we present the proposed CLAWS Net+.
Various components of the model are shown in Fig. \ref{fig:architecture} and discussed below:
\subsection{Training Data Organization Block}
In this block, the input videos are arranged as segments, features of which are extracted using a pre-trained feature extractor. These features are then arranged as batches and a randomly selected batch is forwarded to the backbone network using the random batch selector. These steps are shown in Fig. \ref{fig:architecture} (a)-(e).

\subsubsection{{Video Segment Formation}} {Given a training dataset of $n$ videos,} each video $V_i$ is partitioned into {$m_i$} non-overlapping segments $S_{(i,j)}$ of $p$ frames, where $i \in [1, n]$ is the video index 
and $j \in [1, m_i]$ is the segment index (Fig. \ref{fig:architecture} (a)). The segment size $p$ is kept the same across all training and test videos of a dataset.

\subsubsection{{Feature Extraction}}
For each segment $S_{(i,j)}$, a feature vector $\mathbf{f}_{(i,j)} \in \mathbb{R}^d$ is computed as $\mathbf{f}_{(i,j)}$=$\mathcal{E}(S_{(i, j)})$ using a feature extractor $\mathcal{E}(\cdot)$. In the current work, we use two different feature extractors including C3D proposed by Tran \etal \cite{tran2015c3d} and 3DResNext  proposed by Hara \etal \cite{hara3dcnns}.

\subsubsection{{Batch Formation}}
As shown in Fig. \ref{fig:architecture} (c) \& (d), feature vectors are arranged in non-overlapping batches {$B_k$, each consisting of $b$ consecutive feature vectors such that $B_k=(\mathbf{f}_{(i,j)}, \mathbf{f}_{(i,j+1)}, \cdots, \mathbf{f}_{(i,j+b-1)}) \in \mathbb{R}^{d \times b}$, where $k \in [1, K]$ is the batch index and $K$ is the number of batches in the training data}. All feature vectors within a batch maintain their temporal order, as shown in Fig. \ref{fig:architecture} (d). The proposed batch formation process allows our framework to have more learning instances as each training iteration is carried out using a small portion of a video (batch) instead of a complete video.
For each video we have binary labels as \{normal = 0, abnormal = 1\}. Due to the weakly supervised fashion of training, each batch  inherits the labels of its feature vectors from the parent video.

\subsubsection{{Random Batch Selector}}

In the existing weakly supervised anomaly detection approaches, each training iteration is carried out on one or more complete videos \cite{sultani2018real,zhong2019graph}. In contrast to the existing practice, we propose to extract several batches  from each video. These batches are then input to the backbone network in an arbitrary order using the Random Batch Selector (RBS) (Fig. \ref{fig:architecture} (e)). Such configuration serves the main purpose of minimizing correlation between consecutive batches. 
We observe that breaking the temporal order between consecutive batches results in a significant increase in the backbone performance (see Section \ref{sec:experiments}).

\subsection{Backbone Network Block}
The proposed Backbone Network Block (BNB), shown in Fig. \ref{fig:architecture} (f), contains two fully connected (FC) modules each containing an FC layer followed by a ReLU activation function and a dropout layer. During each iteration, the RBS forwards a random batch to the input layer. 
The output layer is an FC Layer followed by a sigmoid activation function to predict anomaly scores $\mathbf{y} \in \mathbb{R}^b$ in the range $[0, 1]$.
Training of the BNB is carried out using video-level labels. Therefore, an anomalous video batch will have labels $\mathbf{y} = \mathbf{1} \in \mathbb{R}^b$ and a normal video batch will have {labels $\mathbf{y} = \mathbf{0} \in \mathbb{R}^b$, where $\mathbf{1}$ is an all-ones vector, $\mathbf{0}$ is an all-zeros vector, and $b$ is the batch size.}

For the training of our model, three losses including regression loss, temporal smoothness loss and temporal consistency loss are applied directly to the output of BNB at each iteration. Each of these losses are discussed below:

\noindent\textbf{Regression Loss:} The proposed CLAWS Net+ mainly minimizes the mean square error in a batch for each feature label directly inherited from the video-level labels as:
\begin{align}
    \mathcal{L}_{reg} = \frac{1}{b}\sum_{l=1}^{b} (y_l - \hat{y}_l)^2,
    \label{eq:reg}
\end{align}
where $b$ is the batch size, $y_l$ and $\hat{y}_l$ denote the $l$-th ground truth and the predicted label in a batch respectively.

\noindent\textbf{Temporal Smoothness Loss:} Since videos are inherently temporally consistent, a temporal consistency constraint on the predicted segment labels may improve the overall system performance.
In our proposed training data organization block, as the feature vectors in each batch are in temporal order, we apply temporal smoothness loss ($\mathcal{L}_{ts}$) as:

\begin{align}
  \mathcal{L}_{ts} = \frac{1}{b-1}\sum_{l=1}^{b-1} (\hat{y}_{l+1}-\hat{y}_{l})^2,
\label{eq:ts}
\end{align}
where $\hat{y}_l$ is the $l$-th predicted label in a batch of size $b$ feature vectors. 

\noindent\textbf{Sparsity Loss:}
\revised{Since the occurrence of anomalous events is less frequent as compared to the normal events, enforcing a sparsity constraint on the predicted anomaly labels may improve the overall performance of the system.} For this purpose, we enforce the cumulative anomaly score of a complete video to be comparatively small. This loss is computed on each batch during training as:
\begin{align}
 \mathcal{L}_{s} = \frac{1}{b}\sum_{l=1}^{b} \hat{y}_l,
\label{eq:spar}
\end{align}

\subsection{Normalcy Suppression Block}
\label{normalcy_suppression}
Our proposed Normalcy Suppression Block (NSB) consists of multiple normalcy suppression modules (NSMs), as shown in Figs. \ref{fig:architecture} (g)-(h).
Each NSM consists of an FC layer and a softmax layer. 
An NSM serves as a global information collector over a complete batch by computing probability values across temporal dimension.
Based on the number of feature vectors $b$ in each input batch and the FC layer dimension $z$, an NSM estimates a probability matrix $\mathcal{P}$ of size $b \times z$, such that the sum of each column in this matrix is 1. In order to suppress the normalcy within the input batch, an element-wise multiplication between the output of the FC layer and $\mathcal{P}$ is then performed in the corresponding FC Module. The proposed normalcy suppression approach exploits the fact that all normal labeled videos have noise-free labels at segment level and the anomalies do not appear in these videos.

During training, if an input batch is extracted from a normal labeled video, all the features in this batch are labeled as normal and the BNB aims to produce low anomaly scores across the whole batch. Therefore, each NSM learns to minimize the anomaly scores further by distributing its probabilities across the whole input batch, thus complementing the corresponding FC module and avoiding to highlight any part of the input. Such phenomenon fulfills the desideratum behind selecting an element-wise configuration that is to impart more freedom to the NSMs towards minimizing their values for each feature dimension across the whole batch.

In contrast, for an input batch taken from an anomalous video where each feature is labeled as anomalous, one may  expect that the BNB should produce high anomaly scores across the whole batch. However, with normal labeled batches as part of the training data, the BNB is trained to be able to produce low anomaly scores on the normal segments of an anomalous batch. 
Therefore, given the bottleneck imposed by softmax in terms of probability, the NSB learns to suppress the portions of the input batch that do not contribute strongly towards the anomaly scoring in the BNB, consequently highlighting anomalous portions. Thus, it complements the BNB further by suppressing normal segments of an anomalous batch. The NSB is also discussed in Sections \ref{sec:ablation}, \ref{sec:design_choices} \& \ref{sec:qualitative_analysis} where different visualizations and analysis are provided about its working, importance and efficacy.

\begin{figure}[t]
\begin{center}
   \includegraphics[width=1\linewidth]{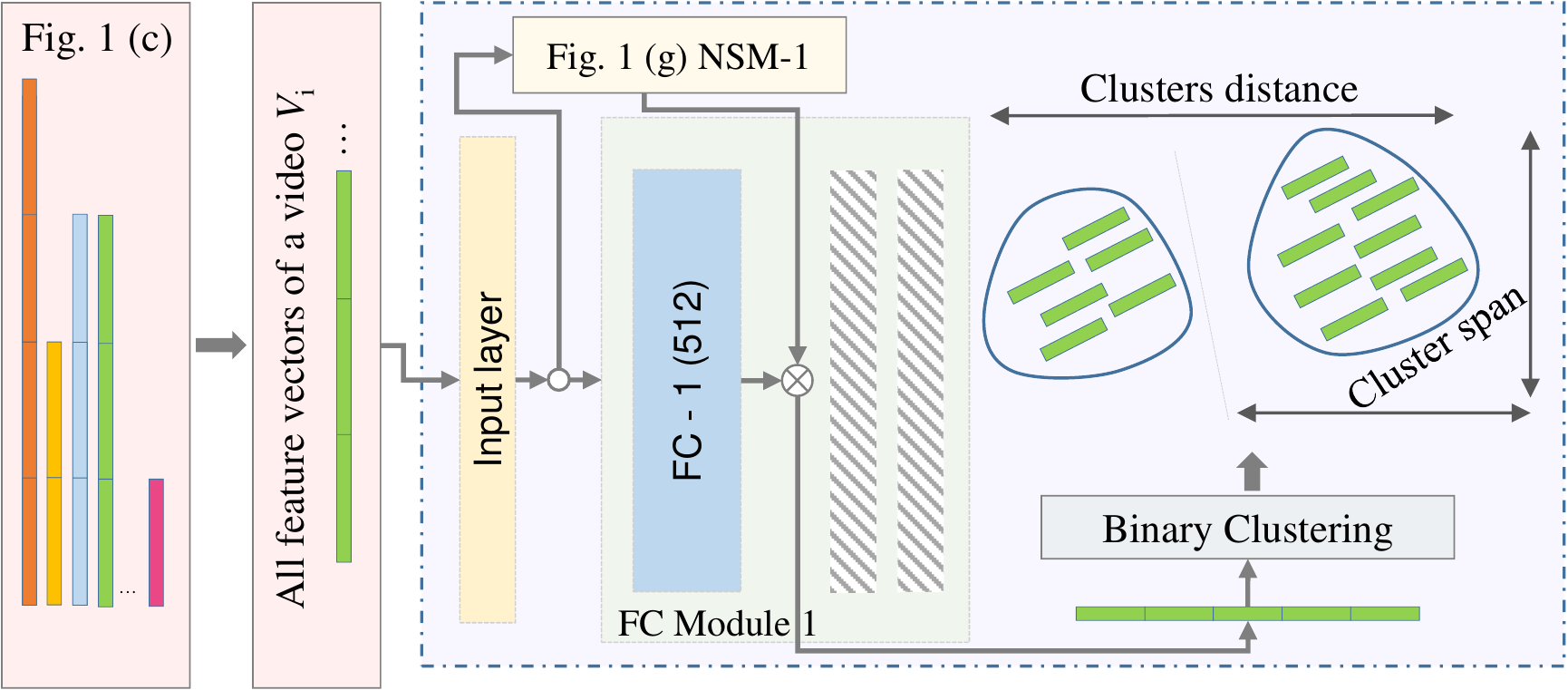}
\end{center}
   \caption{Clustering Loss Block (CLB): intermediate feature representations from a complete video are divided into two clusters in an unsupervised fashion to compute the clustering loss. This loss helps the backbone network to learn more discriminative feature representations for normal and anomalous events. 
    }
\label{fig:CLM}
\end{figure}

\begin{figure}[t]
\begin{center}
   \includegraphics[width=.85\linewidth]{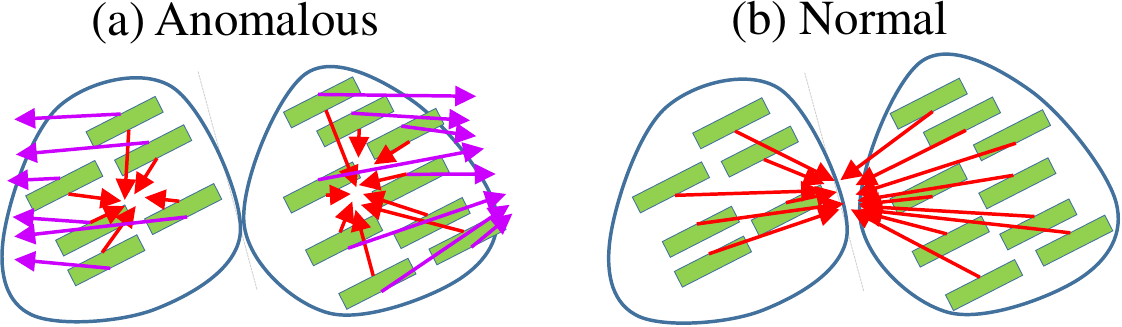}
\end{center}
   \caption{Clustering loss serves two purposes. First, it encourages the backbone network to produce closer clusters in case of a normal video and distant clusters in case of an anomalous video. Second, it encourages the network to produce more compact clusters.}
\label{fig:Cluslosses}
\end{figure}

\subsection{Clustering Loss Block}
The Clustering Loss Block (CLB) is designed to encourage the BNB in learning more discriminative features for the normal and anomalous classes. 
As previously discussed, each feature vector inherits its label, either anomalous or normal, from the parent video. We assume that a normal labeled video has only normal segments while an anomalous labeled video may contain normal segments along with some anomalous segments.
To handle the mislabeled segments in an anomalous video, we propose to cluster the intermediate representations of all the segments in each training video into two clusters. Based on these clusters, we formulate a clustering loss which incorporates clustering distance and compactness, as discussed below:

\noindent\textbf{Clustering Loss:} 
Considering anomaly detection to be a binary problem, two clusters from a normal video should be closer to each other compared to the two clusters from an anomalous video. Therefore, in case of an anomalous labeled video, we enforce to push the centroids of the two clusters away from each other assuming that one cluster should contain normal while the other should contain abnormal segments (Fig. \ref{fig:Cluslosses} (a)). Whereas, in case of a normal labeled video, we enforce to bring the centroids of the two clusters closer (Fig. \ref{fig:Cluslosses} (b)). In addition, we also enforce the clusters to be compact by bringing all elements of the clusters closer to their respective centroids.


In the beginning of each epoch, we compute two clusters, $C^{1}_{i}$ and $C^{2}_{i}$, by using the unit-normalized intermediate representations of all the segments of a video $V_i$. The intermediate representations are extracted from the FC-1 layer of the backbone network block and the clustering is performed with K-means algorithm. Let the centroids of $C^{1}_{i}$ and $C^{2}_{i}$ be  $\mathbf{c}^\mathbf{{1}}_{i}$ and $\mathbf{c}^{\mathbf{2}}_{i}$, then at each training iteration, we calculate the clustering loss for an input batch $B_k$ as:

\begin{equation}
    \mathcal{L}_{c} = 
    \begin{cases}
    \mathcal{L}^N_{c}, \text{\quad if } \mathbf{y}  =  \mathbf{0}\\
    \mathcal{L}^A_{c}, \text{\quad if } \mathbf{y}  =  \mathbf{1},
    \end{cases}
\label{eq:cluster_loss}
\end{equation}
where $\mathbf{y} = \mathbf{0}$ denotes that the batch is taken from a normal video and $\mathbf{y} = \mathbf{1}$ means that the batch is taken from an anomalous video. In either case, we utilize cosine similarity to compute our losses.
Given two vectors, $\mathbf{v}_1$ and $\mathbf{v}_2$, their similarity is given by:
\begin{equation}
    \text{sim}(\mathbf{v}_1, \mathbf{v}_2) = \frac{
    \mathbf{v}_1^\top  \mathbf{v}_2
    }{||\mathbf{v}_1||_2 \cdot || \mathbf{v}_2||_2}.
\end{equation}

For the case of a batch coming from a normal video, the clustering loss $\mathcal{L}^N_{c}$ attempts to maximize the cosine similarity between each segment of the batch and the mean of the centroids $\mathbf{\bar{c}}_i = (\mathbf{c}^\mathbf{1}_{i}+ \mathbf{c}^\mathbf{2}_{i})/2$, defined as:
\begin{equation}
    \mathcal{L}_{c}^N = \frac{1}{b}\sum_{l=1}^{b} ( 1 - \text{sim}(\mathbf{g}_l,\mathbf{\bar{c}}_{i})),
\end{equation}
where $g_l$ is the unit-normalized intermediate representation of a segment in the input batch.
This configuration inherently introduces the minimization of cluster spread and inter-cluster distance as it attempts to converge all elements to the mean of the centroids (Fig. \ref{fig:Cluslosses} (b)). 
In case of an anomalous labeled video, we explicitly aim to increase inter-cluster distance and cluster compactness. Thus, for a batch taken from an anomalous video, the clustering loss 
$\mathcal{L}_{c}^A$ of Eq. \eqref{eq:cluster_loss} is defined as the sum of clustering compactness loss  $\mathcal{L}_{cc}$ and clustering distance loss  $\mathcal{L}_{cd}$:

\begin{equation}
    \mathcal{L}_{c}^A = \alpha \mathcal{L}_{cc} + (1-\alpha)\mathcal{L}_{cd}.
    \label{eq:lca}
\end{equation}
The clustering compactness loss $\mathcal{L}_{cc}$ encourages the network to produce such intermediate representations that can form clusters with smaller spread thus reducing intra-cluster variation:

\begin{equation}
    \mathcal{L}_{cc} = \frac{1}{b_1}\sum_{l=1}^{b_1} (1 - \text{sim}(\mathbf{g}_l^\mathbf{1}, \mathbf{c}^\mathbf{1}_{i}) ) + \frac{1}{b_2}\sum_{l=1}^{b_2} (1 - \text{sim}(\mathbf{g}_l^\mathbf{2}, \mathbf{c}^\mathbf{2}_{i})),
    \label{eq:lcc}
\end{equation}
where feature vectors $(\mathbf{g}^\mathbf{1}_l)_{l=1}^{b_1}$ are taken from $C^1_i$ and $(\mathbf{g}_l^\mathbf{2})_{l=1}^{b_2}$ are taken from $C^2_i$. Furthermore, the clustering distance loss $\mathcal{L}_{cd}$ encourages the network to have intermediate feature vectors that can produce two distinct clusters:
\begin{equation}
    \mathcal{L}_{cd} = \frac{1}{b_1}\sum_{l=1}^{b_1} (1 + \text{sim}(\mathbf{g}_l^\mathbf{1}, \mathbf{c}^\mathbf{2}_{i}) ) + \frac{1}{b_2}\sum_{l=1}^{b_2} (1 + \text{sim}(\mathbf{g}_l^\mathbf{2}. \mathbf{c}^\mathbf{1}_{i})),
    \label{eq:lcd}
\end{equation}
To ensure the inclusion of only high confidence members of the clusters, for both $\mathcal{L}_{cc}$ and $\mathcal{L}_{cd}$, only the feature vectors satisfying the condition of $1-\text{sim}(\mathbf{g}^\mathbf{1}, \mathbf{c}^\mathbf{1}_i) < \beta (1 - \text{sim}(\mathbf{c}^\mathbf{1}_i, \mathbf{c}^\mathbf{2}_i))$ and $1-\text{sim}(\mathbf{g}^\mathbf{2}, \mathbf{c}^\mathbf{2}_i) < \beta (1 - \text{sim}(\mathbf{c}^\mathbf{1}_i, \mathbf{c}^\mathbf{2}_i))$ are included in $\{\mathbf{g}_l^\mathbf{1}\}_{l=1}^{b_1}$ and $\{\mathbf{g}_l^\mathbf{2}\}_{l=1}^{b_2}$ respectively, where $\beta$ is the confidence threshold. Overall, as the clustering is performed on the intermediate representations inferred from the BNB, training on these losses results in an improved capability of our network to represent anomalies and consequently an enhanced anomaly detection performance of the proposed model.

\revised{It should be noted that instead of K-means, several other clustering choices are possible in order to achieve the goal of weakly-supervised training. However, the clustering losses defined by Eqs.~\eqref{eq:lca}-\eqref{eq:lcd} assume clusters to be unimodal in the Euclidean space.  
Therefore, K-means clustering is a prime candidate. 
Also, it is still highly popular among recent methods \cite{asano2019self,caron2019unsupervised,caron2020unsupervised,caron2018deep,alwassel2020self,chen2021multimodal,tian2021divide}.}

\subsection{Training}
CLAWS Net+ exploits video-level annotations as a weak supervisory signal to learn the discrimination between normal and anomalous frames. The overall training of our model is carried out to minimize regression loss (Eq. \eqref{eq:reg}), temporal smoothness loss (Eq. \eqref{eq:ts}), sparsity loss (Eq. \eqref{eq:spar}), and clustering loss (Eq. \eqref{eq:cluster_loss}) computed as:
\begin{align}
  \mathcal{L} = \mathcal{L}_{reg} +\lambda_1 (\mathcal{L}_{s} +\mathcal{L}_{ts})+ \lambda_2\mathcal{L}_{c},
  \label{eq:completeloss}
\end{align}
where $\lambda_1$ and $\lambda_2$ are weight balancing parameters. 

\section{Experiments}
\label{sec:experiments}
In order to evaluate the performance of the proposed CLAWS Net+ framework, a wide range of experiments are performed on three video anomaly detection datasets including UCF-Crime \cite{sultani2018real}, ShanghaiTech \cite{shanghaiTech2017}, and UCSD Ped2 \cite{chan2008ucsd}, and compared with the existing state-of-the-art (SOTA) methods \cite{hasan2016anomaly,lu2013abnormal,sultani2018real,zaheer2020self,zaheer2020cleaning,zhong2019graph,zaheer2020claws,adam2008robust,mahadevan2010anomaly,cong2011sparse,he2018anomaly,xu2017detecting}. All results of the existing SOTA reported in the current paper are taken from the works of the original authors.
\subsection{Datasets}
Details of the three datasets used in our experiments are discussed below:

\noindent\textbf{UCF-Crime} consists of 13 different classes of real-world anomalous events captured through CCTV surveillance cameras spanning 128 hours of videos \cite{sultani2018real}. This dataset is inherently complex due to unconstrained environments. The resolution of each video is standardized to 240 $\times$ 320 pixels.
The training split of this dataset consists of 810 anomalous and 800 normal videos, whereas the test split consists of 140 anomalous and 150 normal videos. Normal labeled videos are free from abnormal scenes. However, abnormal labeled videos contain both normal and anomalous frames. For the training split, only video-level binary labels are available which we use for weakly supervised anomaly detection. In order to facilitate the frame-level evaluation of anomaly detection approaches, the test split provides frame-level binary labels.

\noindent\textbf{ShanghaiTech} consists of staged anomalous events captured in a university campus at 13 different locations with varying viewing angles and lighting conditions. It is a medium scale dataset, containing 437 videos totaling $317,398$ frames of $480 \times 856$ pixels resolution. Since this dataset was originally proposed to train one-class classifiers, only normal data was provided for training. It was recently reorganized by Zhong \etal \cite{zhong2019graph} to facilitate evaluation of weakly supervised binary classification algorithms. To this end, normal and anomalous videos are mixed in both the training and the test splits. The new training split consists of 63 anomalous and 175 normal videos whereas, the new testing split consists of 44 anomalous and 155 normal videos. In order to make a fair comparison, we follow the protocol of Zhong \etal both for training and testing.

\noindent\textbf{UCSD Ped2} is a well-studied dataset for anomaly detection in surveillance videos. It consists of 16 training videos totaling 2,550 frames and 12 test videos totaling 2,010 frames. The normal frames are dominated by pedestrians whereas bicycles, vehicles, skateboards, etc. are considered as anomalies. 
Anomaly examples are not included in the originally proposed training split for this dataset. Therefore, in order to enable weakly supervised binary training, normal and anomalous videos are randomly mixed to create training and test splits as used by the existing SOTA methods \cite{he2018anomaly,zaheer2020self}.

\subsection{Evaluation Metric}
Due to its popularity in similar tasks \cite{lu2013abnormal,zhong2019graph,sultani2018real,hasan2016anomaly}, we use Area Under the Curve (AUC) of the Receiver Operating Characteristic (ROC) computed based on the frame-level annotations of the test videos as our evaluation metric. Usually, a larger AUC means better performance of the system at various thresholds. Moreover, in real-world applications, false alarms of an anomaly detection system should be lower. Therefore, following \cite{sultani2018real,zhong2019graph}, we also report False Alarm Rate (FAR) of our framework. A lower FAR corresponds to lesser number of false alarms.


\begin{figure}[t]
\begin{center}
   \includegraphics[width=0.85\linewidth]{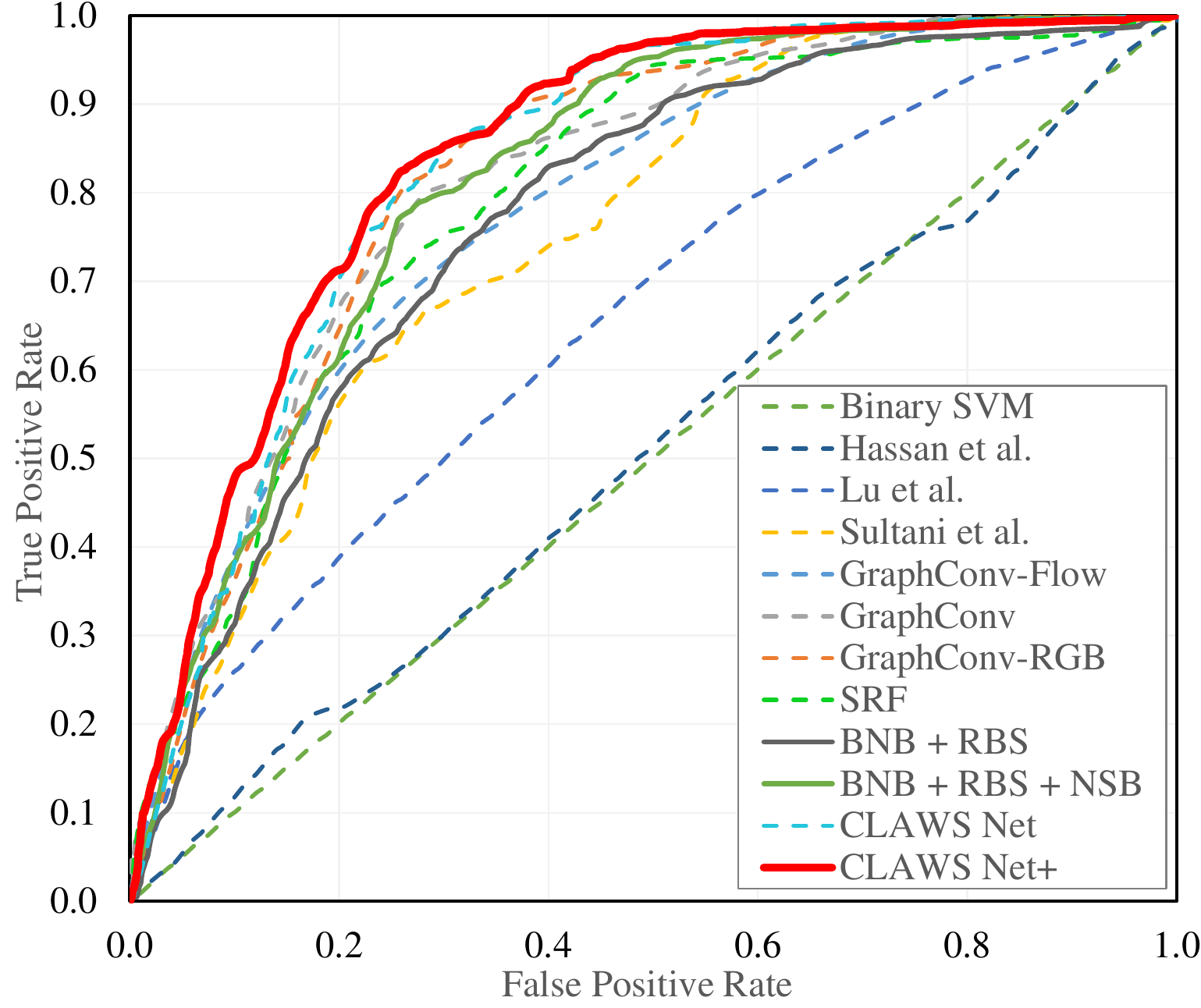}
      \caption[width=0.5\linewidth]{ROC curve comparison of the proposed approach with existing SOTA on UCF-Crime Dataset. }
\label{fig:ROC_plots}
\end{center}
\end{figure}

\subsection{Implementation Details}
Our model is implemented using PyTorch \cite{paszke2017pytorch}. The training is carried out for a total of 100k iterations using RMSProp \cite{tieleman2012lectureRMSProp} optimizer. The initial learning rate is set to 0.0001 and decreased by a factor of 10 after 80k iterations. In all our experiments, $\alpha$ in Eq. \eqref{eq:lca} is set to $0.7$ and the confidence threshold $\beta$ for Eq. \eqref{eq:lcc} \& \eqref{eq:lcd} is set to $0.25$. Furthermore, $\lambda_1$ and $\lambda_2$ in Eq. \eqref{eq:completeloss} are set to $8.0\times10^{-5}$. FC Module 1 and FC Module 2 in the BNB are set to have 512 and 32 channels. Two feature extractors, including C3D \cite{tran2015c3d} and 3DResNext \cite{hara3dcnns}, are used in our experiments. Default feature extraction settings of both architectures are used with segment $p$ set to $16$ frames. In our experiments, we observed that the proposed CLAWS Net+ framework using 3DResNext features performs better than its counterpart using C3D features. Therefore, unless specified otherwise, the default settings of the results reported in this paper consist of 3DResNext features. 

\begin{table}[t]
\begin{center}
\caption{UCF-Crime dataset: Frame-level AUC performance comparison. Backbone Network Block (BNB), Random Batch Selector (RBS), Normalcy Suppression Block (NSB).
}
\resizebox{\columnwidth}{!}{
\begin{tabular}{c|c|c|c} 
\text{\textbf{Method}} & \text{\textbf{Features Type}} & \text{\textbf{AUC\%}} & \text{\textbf{FAR}} \\ \hline
SVM Baseline   & C3D          & 50.00 & - \\ \hline
Hasan et al. \cite{hasan2016anomaly}    & C3D                      & 50.60 & 27.2\\ \hline
Lu et al. \cite{lu2013abnormal}     & C3D                     & 65.51 & 3.10\\ \hline
Sultani et al. \cite{sultani2018real}   & C3D                       & 75.41 & 1.90\\ \hline
Noise Cleaner \cite{zaheer2020cleaning}   & C3D                       & 78.27 & -\\ \hline
SRF \cite{zaheer2020self}   & C3D                       & 79.54 & 0.13\\ \hline
GraphConv\cite{zhong2019graph} &    C3D  &  81.08 & 2.8\\ \hline
GraphConv-Flow  \cite{zhong2019graph} &    $TSN^{OptFlw}$  &  78.08 & 1.10\\ \hline
GraphConv-RGB \cite{zhong2019graph} &    $TSN^{RGB}$  &  82.12 & 0.10\\ \hline
CLAWS Net \cite{zaheer2020claws}  &    $C3D$  &  83.03 & 0.12\\ \hline\hline
BNB  &    C3D &  {69.50} & -\\ \hline
BNB  &    3DResNext &  {68.96} & -\\ \hline
BNB + RBS  &    C3D &  {75.95} & -\\ \hline
BNB + RBS  &    3DResNext &  {76.31} & -\\ \hline
BNB + RBS + NSB  &    C3D &  {80.94} & -\\ \hline
BNB + RBS + NSB  &    3DResNext &  {81.27} & -\\ \hline
\textbf{Proposed CLAWS Net+} &    \textbf{{C3D}}& {\textbf{\red{83.37}}}  & \textbf{\red{0.11}}\\ \hline
\textbf{Proposed CLAWS Net+} &    \textbf{{3DResNext}}& {\textbf{\red{84.16}}} & \textbf{\red{0.09}}
\end{tabular}
}
\label{tab:UCF_crime_AUC}
\end{center}
\end{table}

\subsection{Experiments on UCF-Crime Dataset}
The proposed CLAWS Net+ is trained using only video-level labels in the UCF-Crime dataset. The experiment has been repeated for different types of features including C3D \cite{tran2015c3d} and 3DResNext \cite{hara3dcnns}. In our experiments, CLAWS Net+ with 3DResNext features performed slightly better than the counterpart with C3D features yielding an overall AUC of \red{84.16}\% and an FAR of \red{0.09}, as shown in Table \ref{tab:UCF_crime_AUC}. We also report a component-wise analysis of our framework. As seen, our proposed RBS noticeably enhanced the performance of the BNB to 76.31\% which is superior than the performance of many previously reported results including Hasan et al. \cite{hasan2016anomaly}, Lu et al. \cite{lu2013abnormal} and Sultani et al. \cite{sultani2018real}. This demonstrates the importance of reducing the inter-batch correlation in improving the overall performance. It may be noted that, since we do not need to train a deep action classifier, our proposed approach is fairly simple when compared with GraphConv \cite{zhong2019graph}. However, using the similar C3D features, our CLAWS Net+ depicts \red{2.29\%} improved AUC performance. Moreover, using 3DResNext features, our proposed CLAWS Net+ depicts an elevated performance of \red{2.04\%} in the frame level AUC performance, when compared with the top performing model of GraphConv \cite{zhong2019graph}. Compared to its predecessor CLAWS Net, the CLAWS Net+ demonstrates an improvement of \red{1.13\%} and \red{0.34}\% AUC using 3DResNext and C3D features respectively. This improvement can be attributed to the reformulated clustering loss that incorporates cluster distance and cluster compactness.
The ROC curves of the proposed approach and its different variants are shown in Fig. \ref{fig:ROC_plots}. Both the CLAWS Net and the CLAWS Net+ demonstrated noticeably better performance than the compared methods.


\subsection{Experiments on ShanghaiTech Dataset}
The CLAWS Net+ framework is evaluated on the ShanghaiTech dataset following the test/train split proposed by Zhong et al. \cite{zhong2019graph}.
As seen in Table \ref{tab:shanghaitechAUC}, our proposed  framework outperformed GraphConv \cite{zhong2019graph} by a significant margin of \red{7.02\%}. Moreover, when both algorithms are evaluated using  C3D features, the proposed framework achieved \red{13.68\%} improvement in the AUC.  

\begin{figure}[t]
\begin{center}
    \includegraphics[width=0.85\linewidth]{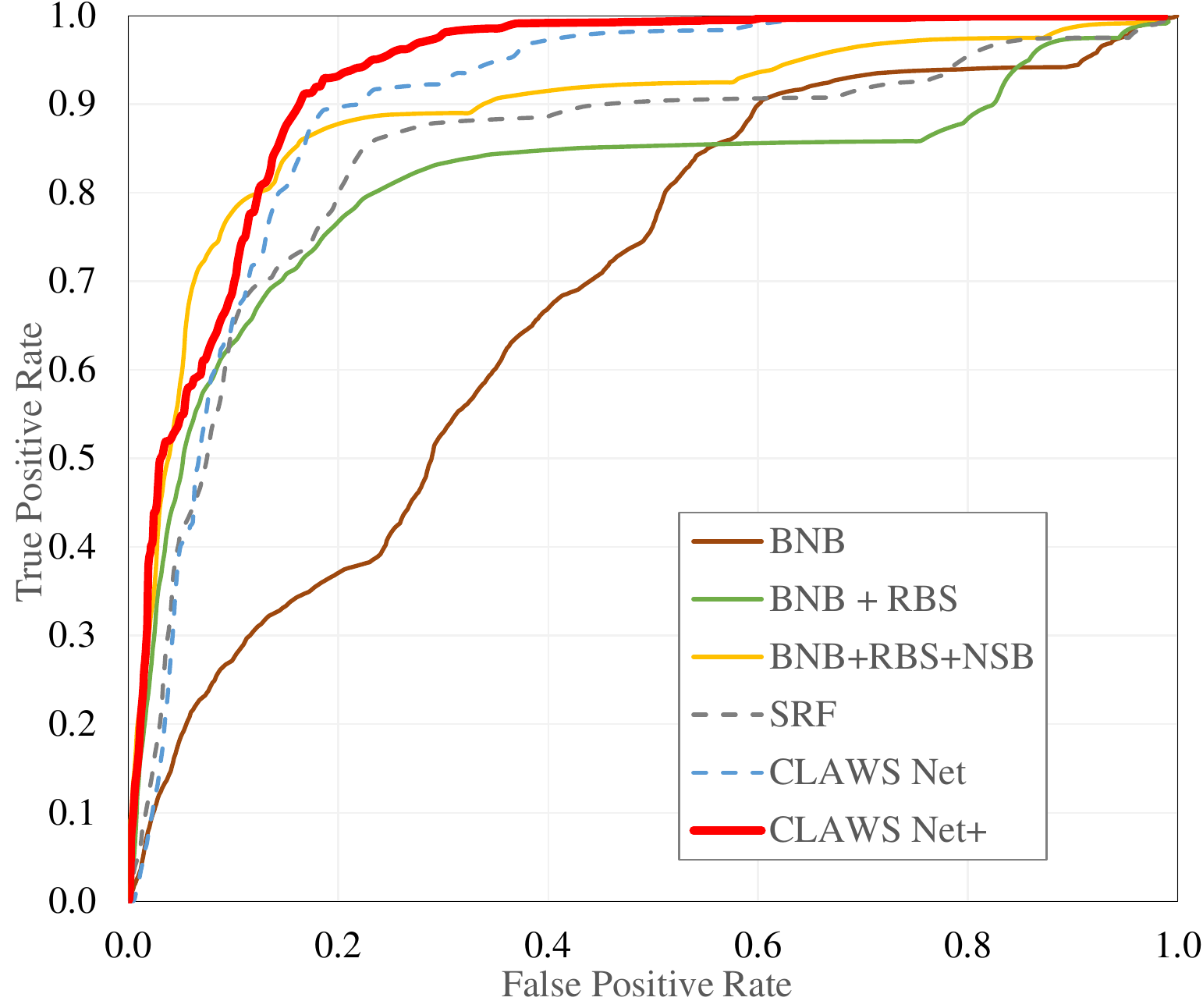}
   \caption{ROC comparison of the variants of our proposed approach on ShanghaiTech dataset.}
    \label{fig:shanghai_tech_roc}
\end{center}
\end{figure}

A performance improvement of 11.7\% is observed by the addition of the RBS to the BNB, using 3DResNext features. In addition to that, when the NSB is added to the BNB+RBS, a performance boost of 8.08\% is observed. Finally the proposed framework, which includes all loss components, further enhances the performance by \red{2.25\%} by achieving \red{91.46\% AUC}. These experiments not only demonstrate the overall robustness of our overall approach but also exhibit the significance of each component.
Fig. \ref{fig:shanghai_tech_roc} shows an ROC curve performance comparison of all variants of the proposed approach with SOTA methods.

\begin{table}[t]
\begin{center}
\caption{ShanghaiTech dataset: frame-level AUC performance comparison of the variants of CLAWS Net+ with existing SOTA methods.}
\resizebox{\columnwidth}{!}{
\begin{tabular}{c|c|c}
\textbf{Method} & \text{\textbf{Features Type}}        & \text{\textbf{AUC(\%)}} \\ \hline
GraphConv \cite{zhong2019graph} &C3D & 76.44  \\ \hline
GraphConv-Flow \cite{zhong2019graph} &$TSN^{Optical Flow}$ & 84.13  \\ \hline
GraphConv-RGB \cite{zhong2019graph} &$TSN^{RGB}$ & 84.44 \\ \hline
Noise Cleaner \cite{zaheer2020cleaning}   & C3D                       & 84.16\\ \hline
SRF \cite{zaheer2020self}   & C3D                       & 84.16\\ \hline
CLAWS Net \cite{zaheer2020claws}   & C3D                       & 89.67\\ \hline\hline

BNB  &    C3D &  {67.50}\\ \hline
BNB  &    3DResNext &  {69.43}\\ \hline
BNB + RBS  &    C3D &  {79.64}\\ \hline
BNB + RBS  &    3DResNext &  {81.13}\\ \hline
BNB + RBS + NSB  &    C3D &  {87.76}\\ \hline
BNB + RBS + NSB  &    3DResNext &  {89.21}\\ \hline
\textbf{Proposed CLAWS Net+} &    \textbf{{C3D}} & \textbf{\red{90.12}}\\ \hline
\textbf{Proposed CLAWS Net+} &    \textbf{3DResNext} & \textbf{\red{91.46}}
\end{tabular}
}
\label{tab:shanghaitechAUC}
\end{center}
\end{table}

\begin{table}[t]
\caption{UCSD Ped2 dataset: frame-level AUC performance comparison of the proposed framework with existing SOTA methods.}
\begin{center}
\resizebox{0.9\columnwidth}{!}{
\begin{tabular}{c|c|c}
\textbf{Method} & \textbf{Features Type}  & \textbf{AUC($\%$)}\\ \hline
Adam \etal \cite{adam2008robust} & - & 63.0  \\ \hline
MDT \cite{mahadevan2010anomaly} & - & 85.0  \\ \hline
SRC \cite{cong2011sparse} & - & 86.1  \\ \hline
AL \cite{he2018anomaly} & - & 90.1  \\ \hline
AMDN \cite{xu2017detecting} & - & 90.8  \\ \hline
GraphConv \cite{zhong2019graph} & TSN$^{OpticalFlow}$ & 92.80\\ \hline
SRF \cite{zaheer2020self} & C3D & 94.47\\ \hline\hline
\textbf{Proposed CLAWS Net+} & \textbf{C3D} & \textbf{\red{94.91}}\\ \hline
\textbf{Proposed CLAWS Net+} & \textbf{3DResNext} & \textbf{\red{95.79}}
\end{tabular}
}
\end{center}
\label{tab:ped2}
\end{table}

\begin{table}[b]
\begin{center}
\caption{Ablation analysis of the proposed framework on UCF-Crime dataset using C3D Features. Backbone Network Block (BNB), Random Batch Selector (RBS), Normalcy Suppression Module (NSM), Sparsity loss ($\mathcal{L}_s$) , Temporal Smoothness loss ($\mathcal{L}_{ts}$), Clustering loss ($\mathcal{L}_c$).}
\resizebox{1\linewidth}{!}{
\begin{tabular}{|c|c|c|c|c|c|c|}
\hline
\textbf{BNB} & \textbf{RBS} & \textbf{NSM-1} & \textbf{NSM-2} & \bm{$\mathcal{L}_s + \mathcal{L}_{ts}$} & \bm{$\mathcal{L}_{c}$} &\textbf{AUC (\%)} \\ \hline\hline
\multicolumn{7}{|c|}{\textbf{Top-down Approach}}                    \\ \hline\hline
\ding{51}   & -   & \ding{51}     & \ding{51}     & \ding{51}        &  \ding{51}   & \red{80.14}    \\ \hline
\ding{51}   & \ding{51}   & -     & \ding{51}     & \ding{51}        &  \ding{51}   & \red{76.82}    \\ \hline
\ding{51}   & \ding{51}   & \ding{51}     & -     & \ding{51}        &  \ding{51}   & \red{80.13}    \\ \hline
\ding{51}   & \ding{51}   & -     & -     & \ding{51}         & \ding{51}   & \red{76.58}    \\ \hline
\ding{51}   & \ding{51}   & \ding{51}     & \ding{51}     & -          & \ding{51}   & \red{82.57}    \\ \hline\hline
\multicolumn{7}{|c|}{\textbf{Bottom-up Approach}}                   \\ \hline\hline
\ding{51}   & -   & -     & -     & -        & -     & {69.50}    \\ \hline
\ding{51}   & \ding{51}   & -     & -     & -        & -  & {75.95}    \\ \hline
\ding{51}   & \ding{51}   & \ding{51}     & -     & -        & -    & {78.60}    \\ \hline
\ding{51}   & \ding{51}   & \ding{51}     & \ding{51}     & -        & -   & {80.94}    \\ \hline
\ding{51}   & \ding{51}   & \ding{51}     & \ding{51}     & \ding{51}        & -  & {81.53}    \\  \hline\hline
\ding{51}   & \ding{51}   & \ding{51}     & \ding{51}     & \ding{51}        & \ding{51}    & \red{83.37}   \\ \hline
\end{tabular}
}

\label{tab:ablation}
\end{center}
\end{table}

\subsection{Experiments on USCD Ped2}

The proposed framework is evaluated on the Ped2 dataset following the evaluation protocol used in the existing SOTA methods \cite{he2018anomaly,zaheer2020self}. In a series of experiments, 6 anomalous and 4 normal videos are randomly selected for training and the remaining videos are included in the test set. Average AUC performance is reported over five-fold experiments. Our framework demonstrates \red{95.79}\% AUC performance, which is superior than the existing approaches \cite{adam2008robust,mahadevan2010anomaly,cong2011sparse,he2018anomaly,xu2017detecting,zhong2019graph,zaheer2020self}.

\begin{table}[t]
\caption{Area Under the Curve (AUC) comparison of various normalcy suppression (NS) configurations (shown in Fig. \ref{fig:suppression_types}) on UCF Crime dataset using 3DResNext features.}
\begin{center} 
\begin{tabular}{|l|l|}
\hline
\textbf{Normalcy suppression}  & \textbf{AUC \%} \\ \hline
Element-wise Temporal NS (CLAWS Net+)  & \red{84.16}  \\ \hline
Feature Vector NS & 81.79  \\ \hline
\begin{tabular}[c]{@{}l@{}}Element-wise Spatial NS\end{tabular} & 78.39  \\ \hline
\begin{tabular}[c]{@{}l@{}}CLAWS Net+ without any NSM \end{tabular} & 77.11  \\ \hline
\end{tabular}
\end{center}
\label{tab:suppression_types}
\end{table}

\subsection{Ablation}
\label{sec:ablation}
In addition to the comparisons provided between different variants of our proposed approach in Tables \ref{tab:UCF_crime_AUC} \& \ref{tab:shanghaitechAUC}, 
two types of detailed ablation studies i.e., top-down and bottom-up, are performed on our model trained on UCF-Crime dataset (see Table \ref{tab:ablation}). The top-down approach is started with the whole framework of CLAWS Net+ and individual modules are removed to observe the performance degradation. The bottom-up is started with the Backbone Network Block (BNB) and different blocks are then added to observe the performance improvement. To limit the extent of the experiments, we report ablation using only C3D features.


\begin{figure*}[t]
\begin{center}
  \includegraphics[width=.85\linewidth]{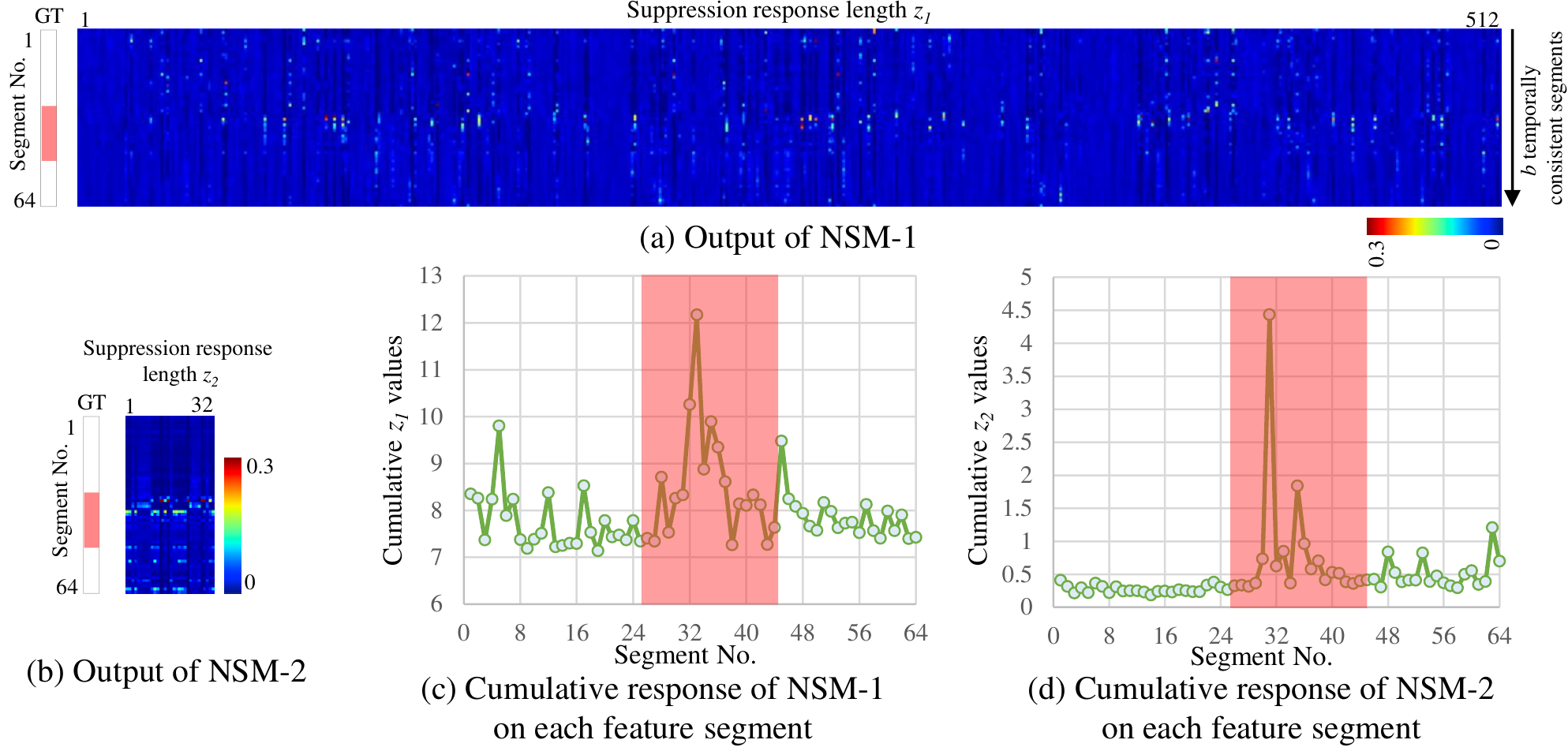}
\end{center}
  \caption{Visualizations of the Normalcy Suppression Modules outputs: (a) Probability map generated by NSM-1 and (b) NSM-2, on an anomalous video batch. Both modules have learned to suppress normal input regions, consequently highlighting the anomalous regions.  (c) \& (d) Segment-wise cumulative probability scores are high in the anomalous regions. Anomaly ground-truth is shown as red colored rectangles.  } 
\label{fig:attention_batches}
\end{figure*}

\begin{figure*}[t]
\begin{center}
   \includegraphics[width=1\linewidth]{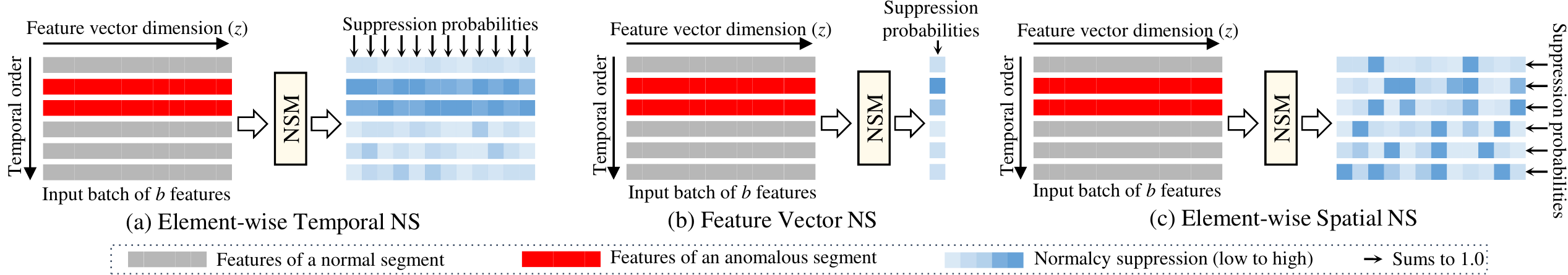}
\end{center}
   \caption{Visualization of the three types of Normalcy Suppression (NS) configurations investigated. (a) Element-wise NS (as proposed in the CLAWS Net+). (b) Feature Vector NS: Suppression Probabilities (Softmax) are calculated along temporal dimension considering one feature vector at a time. (c) Element-wise Spatial NS: Suppression Probabilities (Softmax) are calculated along each feature vector.
}
\label{fig:suppression_types}
\end{figure*}

In the top-down approach, a drop of \red{3.23\%} AUC is observed with the deletion of the RBS from the proposed CLAWS Net+ framework. Compared to this, in the bottom-up approach, the addition of the RBS to the BNB yielded an improvement of \red{6.45\%}. This demonstrates that although a part of the performance of the RBS may be compensated by other modules in the complete framework, it cannot be fully replaced. 

Deletion of the NSB (NSM-1 \& NSM-2) from the overall framework dropped the performance by \red{6.79}\% while its addition to BNB+RBS in the bottom-up approach resulted in an improvement of \red{4.99\%}. This shows that the effect of the NSB in the top-down approach on the overall system is significantly larger than that of observed in the bottom-up approach which is due to its direct impact on the clustering loss block. In addition to that, NSB also indirectly effects the other losses computed to train our proposed framework.
In order to further explore the role of the NSB, the responses of NSM-1 and NSM-2 for a batch taken from an anomalous test video are visualized in Figs. \ref{fig:attention_batches} (a) \& (b). It may be noticed that the response on normal events is significantly smaller as compared to the response on anomalous events. The output of the NSM layers is applied on the output of the corresponding FC layers, resulting in suppression of the normal regions. Additionally, the cumulative responses of NSM-1 and NSM-2 modules are computed over each segment for visualization, as shown in Figs. \ref{fig:attention_batches} (c) \& (d). The cumulative response in the normal region is significantly lower than the response in the anomalous regions. These visualizations demonstrate that the proposed NSM-1 and NSM-2 modules have successfully learned normalcy suppression which results in highlighting the anomalous regions.

An individual comparison of the importance of the NSM-1 and the NSM-2 modules is also performed (Table \ref{tab:ablation}). Removal of NSM-1 and NSM-2 caused performance degradation of \red{6.55\%} and \red{3.24\%} respectively in the top-down approach. In the bottom-up approach, addition of NSM-1 and NSM-2 caused an improvement of 2.65\% and 2.34\% respectively. Both experiments demonstrate a relatively more significance of NSM-1 compared to NSM-2 which is mainly because of its larger FC layer size. Also, NSM-1 directly effects the features used for clustering loss module as the clustering is performed after applying the NSM-1 response to the FC-1 output (Fig. \ref{fig:CLM}).

Table \ref{tab:ablation} also shows that the addition of the temporal smoothness loss ($\mathcal{L}_{ts}$) and the sparsity loss ($\mathcal{L}_s$) brings the performance to 81.53\%. On the other hand, removal of these losses from the complete CLAWS Net+ framework brings down the performance to \red{82.57\%}. This again demonstrates that although the presence of other components may reduce the relative importance, the optimal performance cannot be achieved without each component. We also observed the performance contribution by the clustering loss block. Removing it from the overall framework caused a degradation of \red{1.84\%} in performance, demonstrating its crucial importance in the overall system.

\begin{figure}[t]
\vspace{-3mm}
\begin{center}
   \includegraphics[width=1\linewidth]{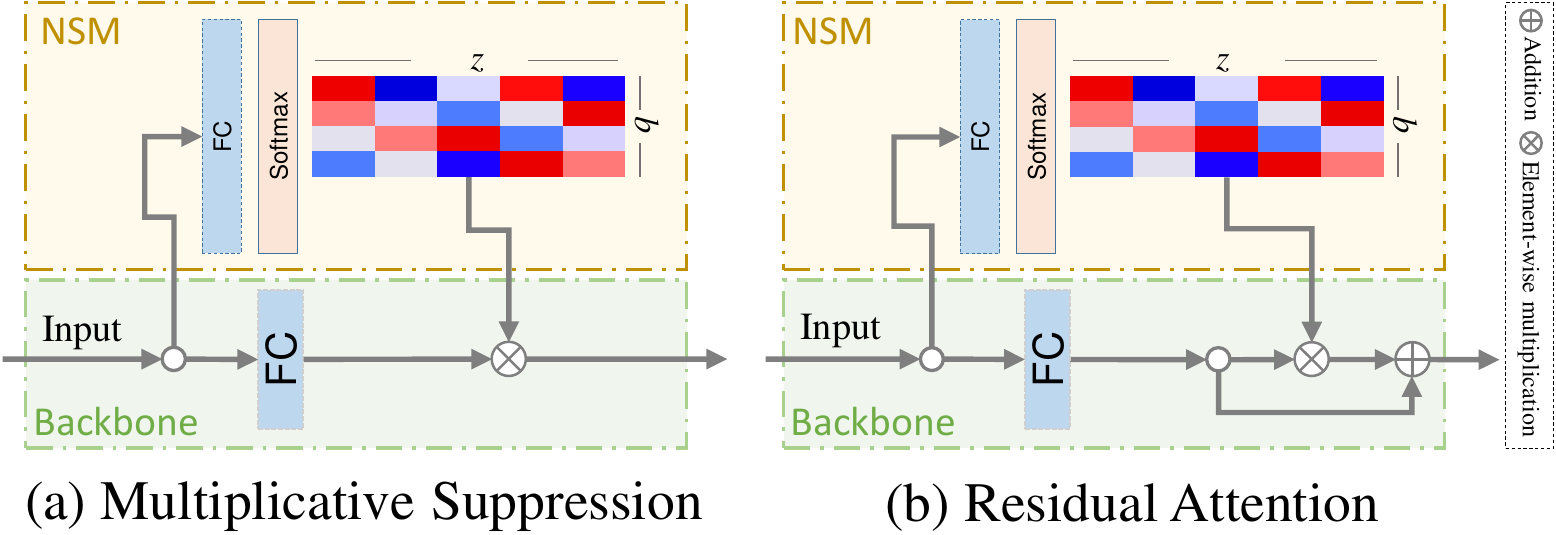}
\end{center}
   \caption{(a) Multiplicative suppression, as utilized in our proposed CLAWS Net+ (b) Residual attention as proposed by Wang \etal \cite{wang2017residual}. }
   \label{fig:attnsupp}
\end{figure}

\begin{table}[t]
\caption{AUC comparison of the proposed multiplicative suppression mechanism with the residual suppression (Fig. \ref{fig:attnsupp}) on UCF-Crime dataset using 3DResNext features. 
}
\begin{center} 
\begin{tabular}{|l|l|}
\hline
\textbf{Normalcy suppression}  & \textbf{AUC \%} \\ \hline
Multiplicative suppression (CLAWS Net+)   & \red{84.16}  \\ \hline
Residual suppression & 78.14 \\ \hline
CLAWS Net+ without any NSM & 77.11 \\ \hline
\end{tabular}
\end{center}
\label{tab:attentionvssuppression}
\end{table}

\begin{figure*}[t]
\begin{center}
   \includegraphics[width=1\linewidth]{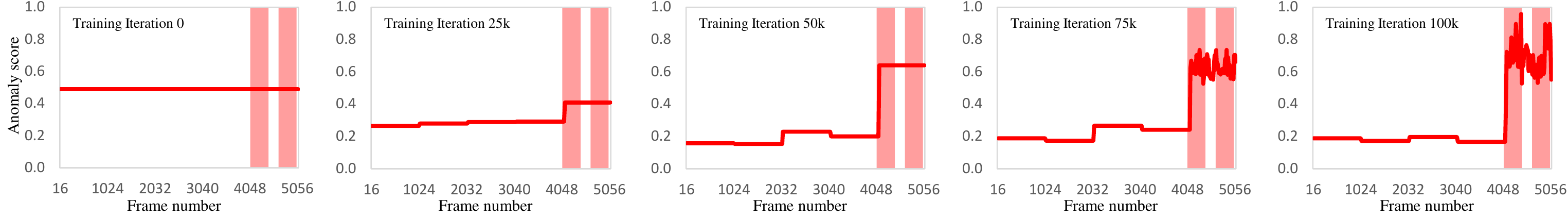}
\end{center}
   \caption{Evolution of the frame-level anomaly scores output by CLAWS Net+, on \textit{Shooting046} video from UCF-Crime, over several training iterations. Although weakly supervised, our framework learns to produce significantly higher scores in the anomalous portions whereas lower scores in the normal portions. Anomaly ground-truth is shown as red colored rectangles.}
\label{fig:scores_evolution}
\end{figure*}

\begin{figure*}[t]
\begin{center}
   \includegraphics[width=.9\linewidth]{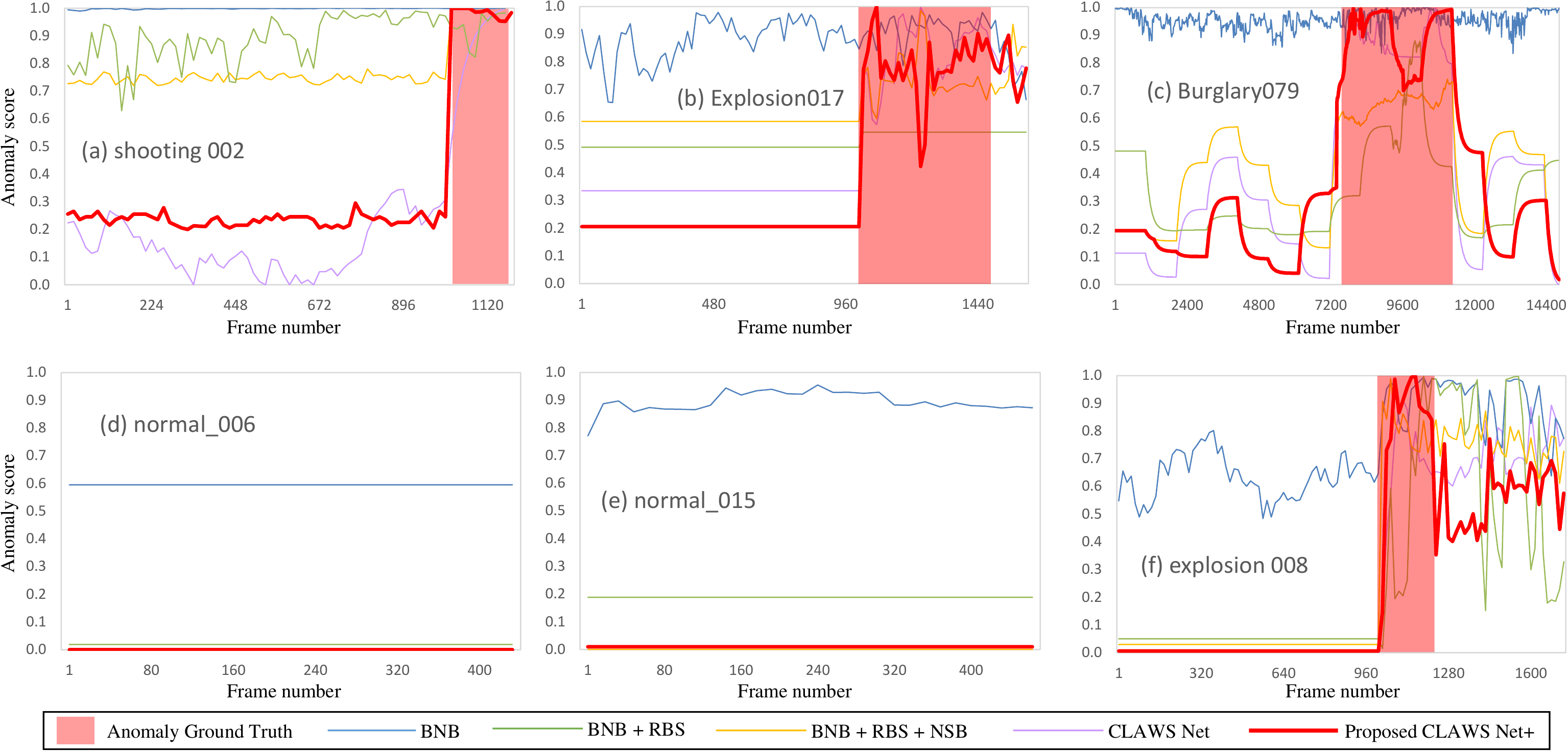}
\end{center}
   \caption{Qualitative comparison of the variants of our proposed approach on several test videos from UCF-Crime. Anomaly cases are shown in (a), (b) \& (c) whereas normal cases are depicted in (d) \& (e). A relatively unsuccessful case is illustrated in (f) in which, the anomaly score remains higher after the explosion. Anomaly ground-truth is shown as red colored rectangles.}
\label{fig:qualitative_results}
\end{figure*}

\begin{figure}[t]
\vspace{-3mm}
\begin{center}
   \includegraphics[width=1\linewidth]{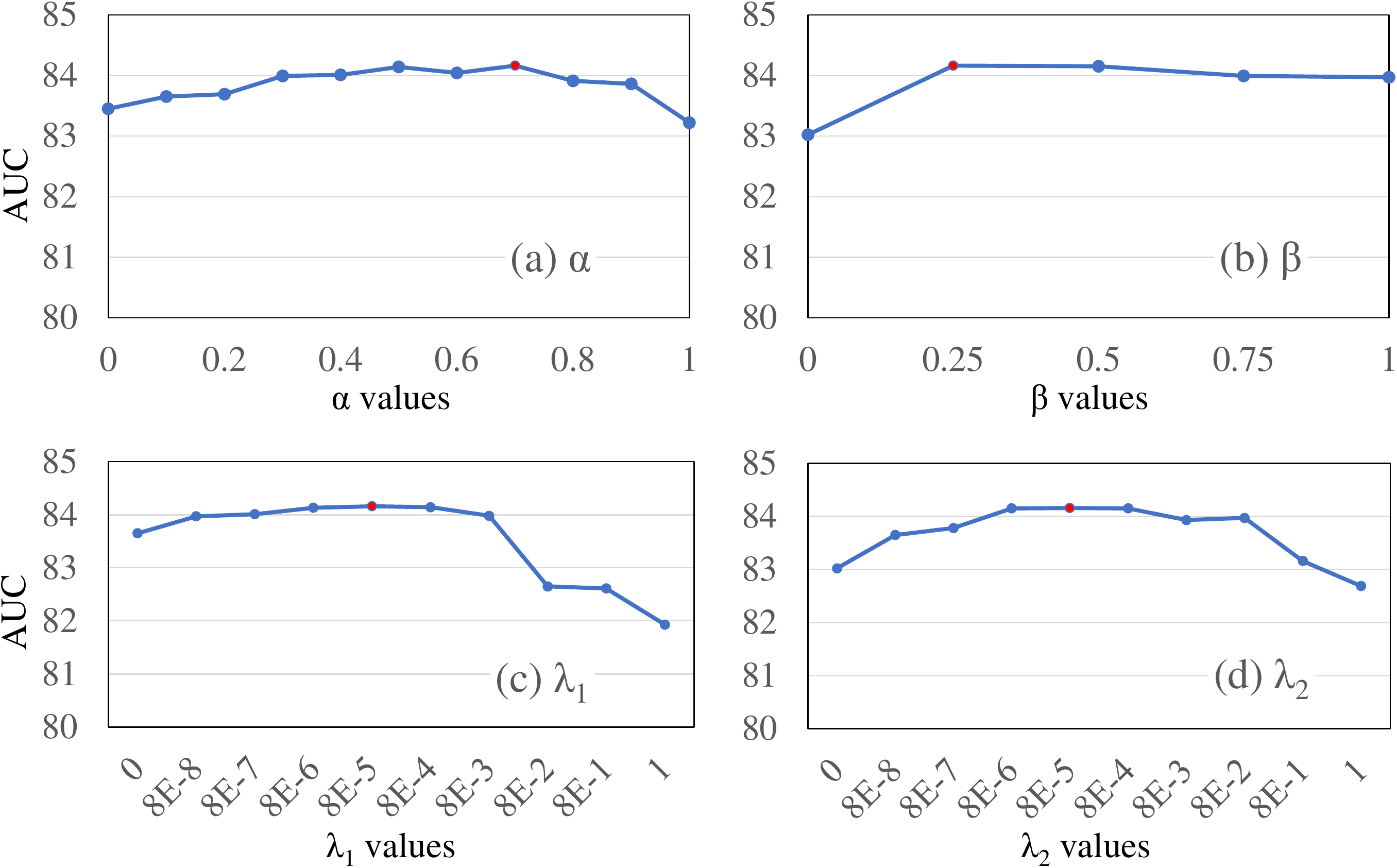}
\end{center}
   \caption{\revised{Empirical analysis of the hyperparameter values used in CLAWS Net+. Red dots in each graph represent the values used in our experiments. It can be seen that our approach retains almost identical performance over a fairly wide range of values demonstrating stability and robustness towards perturbation in these parameters. All experiments are conducted using 3DResNext features extracted from UCF-Crime dataset.}}
   \label{fig:hyperparameters_analysis}
\end{figure}

\subsection{Design Choices}
\label{sec:design_choices}
We analyzed different design choices in our proposed framework and discuss the performance of each choice in this section. 

\subsubsection{Which type of Normalcy Suppression (NS) is better?}
As explained in Section \ref{normalcy_suppression}, the proposed normalcy suppression mechanism calculates the probabilities which are applied on each feature along the temporal axis of the input batch in  an  element-wise  fashion (Fig. \ref{fig:suppression_types} (a)). In the following, we explore two other possible attention mechanisms for the implementation of NS.

\noindent\textbf{Feature Vector NS:}  In this setting,  suppression values are calculated in a segment-wise fashion, i.e., one value is computed for each feature vector within a batch, and the whole feature vector is suppressed based on the computed value (Fig. \ref{fig:suppression_types} (b)). This NSM estimates a probability vector of size $b \times 1$ for each batch, such that the sum of this vector is 1.00.

\noindent\textbf{Element-wise Spatial NS:} In  order  to  provide  a comparison  of  learning probability scores across the temporal dimension in  our proposed NSB (Section \ref{normalcy_suppression}), we also experiment with a normalcy suppression scheme which computes probability scores across the spatial dimension. This setting, referred to as element-wise spatial NS  in Fig. \ref{fig:suppression_types} (c), computes suppression values along each feature vector of the input batch. Given the number of feature vectors $b$ in each input batch and the FC layer dimension $z$, this NSM estimates a probability matrix of size $b \times z$, such that the sum of each row in this matrix is 1.00.

Table \ref{tab:suppression_types} summarizes the frame level AUC performance of these three configurations. It can be seen that the element-wise temporal NS outperforms the other two counterparts with a noticeable margin. It is interesting to observe that the performance of the feature vector NS is relatively closer to the element-wise temporal NS, which is due to the reason that these two are relatively similar in essence. Both of these learn to minimize the effects of normal features towards anomaly scoring. However, the element-wise temporal NS performs it with an additional freedom of operating at each feature dimension. Therefore, both of these suppression mechanisms utilizing temporal properties outperform the third scheme, element-wise spatial NS, significantly.

\subsubsection{Normalcy Suppression or Attention?}
Some existing literature have reported superior performance of  attention mechanisms that highlight the important parts or features in a given input \cite{woo2018cbam,hu2018squeeze,wang2017residual}. In contrast, attributed to the availability of noise-free normal data, our proposed formulation approaches the problem in terms of suppressing features belonging to normal data by utilizing a multiplicative approach. In order to evaluate the efficacy of our approach, we also experiment using another NSM that contains residual attention proposed by Wang \etal \cite{wang2017residual}. In this approach, the actual features are added again to their attended features (Fig. \ref{fig:attnsupp}). Table \ref{tab:attentionvssuppression} shows a comparison of the results between the residual attention and our multiplicative suppression approach, demonstrating the importance of the latter for the proposed weakly supervised anomaly detection framework. Compared with the model without any NSM, the residual attention shows only a slight improvement of \red{1.03\%} whereas the proposed multiplicative suppression shows an improvement of \red{7.05\%}.



\subsection{Qualitative Analysis}
\label{sec:qualitative_analysis}
The response of the proposed anomaly detection framework CLAWS Net+ for the \textit{shooting046} video taken from the UCF-Crime dataset is observed over several training iterations as shown in Fig. \ref{fig:scores_evolution}. The discrimination between the normal and anomalous regions enhances as the training iterations proceed, demonstrating the convergence of the proposed framework towards assigning high values to anomalous regions and low values to normal regions.

A comparison of the final response of CLAWS Net+ and other variants on four anomalous and two normal test videos from the UCF-crime dataset is shown in Fig. \ref{fig:qualitative_results}. In most of these cases, the BNB was not able to discriminate well between the anomalous and normal video content. Addition of the RBS improved the discrimination however, in some cases such as \textit{shooting002}, it remained small. Further addition of the NSB significantly improved the capability of the proposed system for discriminating normal and anomalous events. The proposed normalcy suppression not only pushed the normal events scores towards 0 but also created a smoothing effect. Finally, The response of the complete CLAWS Net+ is more discriminative and stable compared to all other variants in almost all cases. 

One interesting case is shown in Fig. \ref{fig:qualitative_results} (f) for the case of \textit{explosion008} in which the system response remained higher after the ground truth region of the anomalous event is over. Understandably, it is due to the aftermath of an explosion in which the situation remains abnormal for a significant amount of time. Since the dataset annotations only highlight the explosion duration, the high system response after the explosion is considered as false positive. A similar behavior can also be observed in Fig. \ref{fig:qualitative_results} (b) for the case of \textit{explosion017}.




\subsection{\revised{Hyperparameters Analysis}}
\label{sec:hyperparameters_analysis}

\revised{In order to observe the sensitivity of our approach on $\alpha$, $\beta$, $\lambda_1$, and $\lambda_2$ used in Eqs. \eqref{eq:lca}, \eqref{eq:lcc}, \eqref{eq:lcd}, and \eqref{eq:completeloss}, we conducted extensive empirical analysis by varying the values of these parameters on UCF Crime dataset, using 3DResNext features, and report the results in Fig. \ref{fig:hyperparameters_analysis}. As in Eq. \eqref{eq:lca}, $\alpha$ controls the ratio between clustering distance loss and cluster center loss. It is evident from Fig. \ref{fig:hyperparameters_analysis}(a) that our CLAWS Net+ demonstrates consistent performance over a wide range of $\alpha$ values: $0.3\le\alpha\le0.8$. 
For $\beta$ in Eqs. \eqref{eq:lcc} and \eqref{eq:lcd}, intuitively we aim to include the features with high confidence for clustering loss calculation. 
As seen (Fig. \ref{fig:hyperparameters_analysis}(b)), if we set $\beta=0$, the clustering loss is not applied as no feature vector satisfies the condition and the performance drops noticeably. Best performance is observed for $\beta=0.25$ because it includes only high confidence features.
As we further increase the values of $\beta$, a slight drop in performance can be observed due to the inclusion of more low confidence features, however the difference is negligible.
Parameters $\{\lambda_1,\lambda_2\}$ in Eq. \eqref{eq:completeloss} control the contribution of temporal smoothness loss, sparsity loss, and clustering loss.  We  empirically observe (Fig. \ref{fig:hyperparameters_analysis}(c)\&(d)) that  CLAWS Net+ achieves almost identical performance over a  wide range of these parameters: $8\times10^{-8}\le\lambda_1\le8\times10^{-3}$ and $8\times10^{-6} \le \lambda_2 \le 8\times10^{-3}$. For $\lambda_1=0$ performance drops to 83.65\% which is due to the removal of smoothness and sparsity losses. For $\lambda_1=1$, performance drops due to more emphasis on these losses. Similarly, for $\lambda_2=0$, the performance drops due to the removal of clustering loss. Whereas, for high values of this parameter, $\lambda_2>8\times10^{-2}$, the performance drops due to more emphasis on this loss. These experiments show that our proposed CLAWS Net+ is not sensitive to small variations in these parameters and the performance remains stable over a significantly wide range of each parameter.
}

\section{\revised{Conclusions and Future Works}}
In this work, a weakly-supervised anomalous event detection system is proposed which requires only video-level labels during training. In order to train the proposed system, a batch based training is devised which is different from the previously used full-video based training methods. A video may be divided into several batches depending upon its length while a batch consists of several temporally ordered video segments. A random batch selector (RBS) is also proposed to break the inter-batch correlation. The RBS has demonstrated a significant performance gain compared to the backbone network block. A normalcy suppression block is also proposed which suppresses the features corresponding to the normal events in an overall abnormal labeled video by collaborating with the backbone network. Such normalcy suppression has resulted in an improved discrimination between normal and anomalous input regions. Moreover, to improve the representation learning of the anomalous and the normal events, a clustering based loss is formulated, which jointly improves the capability of the proposed system to better discriminate anomalies from normal events. Evaluation of the proposed CLAWS Net+ is performed on three anomaly detection benchmark datasets including UCF-Crime, ShanghaiTech and UCSD Ped2. Comparisons with the existing SOTA methods demonstrate excellent performance of the proposed framework.

\revised{Due to its applicability in weakly-supervised learning, in future, this work can be explored for its applications in different other computer vision domains including video object detection, action localization in untrimmed videos, etc. Moreover, the idea of unsupervised clustering for weakly-supervised training signal can also be explored for other types of data including images and tabular datasets.}






%

\section*{Acknowledgments}
This  work  was supported by the Institute of Information 
\&  communications  Technology  Planning \&  Evaluation (IITP) grant
funded by the  Korea government (MSIT) (No. 2017-0-00306, Development of Multimodal Sensor-based Intelligent Systems for Outdoor Surveillance Robots, 50\%) and (No. 2019-0-01309, 
Development of AI  Technology for  Guidance of a Mobile Robot 
to its  Goal  with  Uncertain Maps in Indoor/Outdoor 
Environments, 50\%) 


\ifCLASSOPTIONcaptionsoff
  \newpage
\fi



%

\bibliographystyle{IEEEtran} 
\bibliography{main}

\vspace{-15mm}
\begin{IEEEbiography}[{\includegraphics[width=1in,height=1.25in,clip,keepaspectratio]{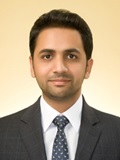}}]{Muhammad Zaigham Zaheer}
is currently a PhD candidate at the University of Science and Technology, Daejeon, Korea. He is also associated with the Electronics and Telecommunications Research Institute, Daejeon, Korea, as a student researcher. Previously, he received his MS degree from Chonnam National University, Gwangju, Korea, in 2017 and his BS degree from Pakistan Institute of Engineering and Applied Sciences, Islamabad, Pakistan, in 2012. His current research interests include computer vision,  anomaly detection in images/videos, semi-supervised/self-supervised learning and video object segmentation.
\end{IEEEbiography}
\vspace{-15mm}
\begin{IEEEbiography}[{\includegraphics[width=1in,height=1.25in,clip,keepaspectratio]{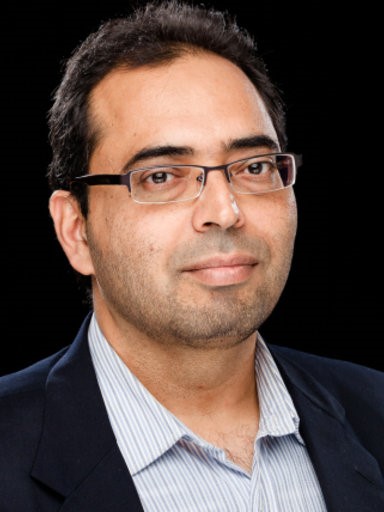}}]{Arif Mahmood} is a Professor and Chairperson of Computer Science Department in Information Technology University and Director Computer Vision Lab. His current research directions in Computer Vision are person pose detection and segmentation, crowd counting and flow detection, background-foreground modeling in complex scenes, object detection, human-object interaction detection and abnormal events detection. He is also actively working in diverse Machine Learning applications including  cancer grading and prognostication using histology images, predictive auto-scaling of services hosted on the cloud and the fog infrastructures, and environmental monitoring using remote sensing. He has also worked as a Research Assistant Professor with the School of Mathematics and Statistics, University of the Western Australia (UWA) where he worked on Complex Network Analysis. Before that he was a Research Assistant Professor with the School of Computer Science and Software Engineering, UWA and performed research on face recognition, object classification and action recognition. 
\end{IEEEbiography}
\vspace{-15mm}
\begin{IEEEbiography}[{\includegraphics[width=1in,height=1.25in,clip,keepaspectratio]{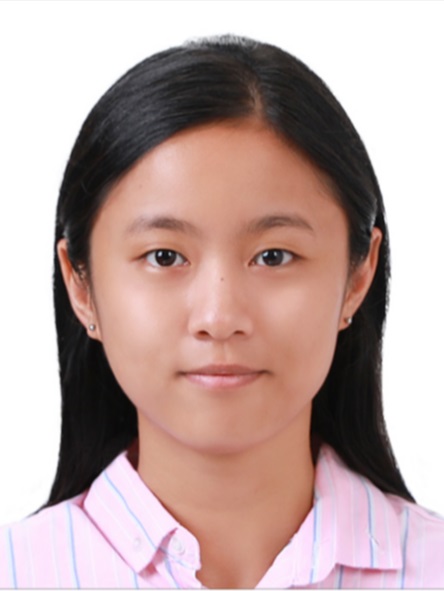}}]{Marcella Astrid}
Marcella Astrid received her BEng in computer engineering from the Multimedia Nusantara University, Tangerang, Indonesia, in 2015, and the MEng in computer software from the University of Science and Technology (UST), Daejeon, Korea, in 2017. At the same university, she is currently working towards her PhD degree in computer science. Her recent interests include deep learning and computer vision.
\end{IEEEbiography}
\vspace{-15mm}
\begin{IEEEbiography}[{\includegraphics[width=1in,height=1.25in,clip,keepaspectratio]{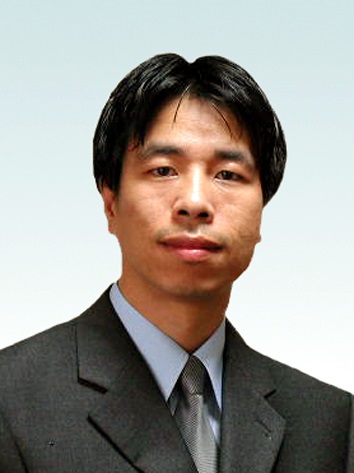}}]{Seung-Ik Lee}
Seung-Ik Lee received his BS, MS, and PhD degrees in computer science from Yonsei University, Seoul, Korea, in 1995, 1997 and 2001, respectively. He is currently working for the Electronics and Telecommunications Research Institute, Daejeon, Korea. Since 2005, he has been with the Department of Computer Software, University of Science and Technology, Daejeon, Korea, where he is a professor. His research interests include  computer vision, deep learning, and reinforcement learning.
\end{IEEEbiography}



\end{document}